\pdfoutput=1
\documentclass[11pt,a4paper]{article}
\usepackage[hyperref]{acl2020}
\usepackage{url}
\usepackage{times}
\usepackage{latexsym}

\usepackage{graphicx}
\usepackage{microtype}
\usepackage{soul}
\usepackage{xspace}
\newcommand{\gerstim}{\textsc{GerSti}\xspace}
\newcommand{\gne}{\textsc{Gne}\xspace}
\newcommand{\F}{F$_1$\xspace}
\newcommand{\crf}{\textsc{CRF}\xspace}
\newcommand{\xlmr}{\textsc{XLM-R}\xspace}
\newcommand{\rt}[1]{\rotatebox{90}{#1}}

\newcommand{\leftbr}{$\left[\right.$\xspace}
\newcommand{\rightbr}{$\left.\right]$\xspace}
\usepackage{booktabs}
\usepackage{tabularx}
\usepackage{multirow}
\usepackage{makecell}

\usepackage{xcolor}
\usepackage{pgfplots}
\usepackage{tikz,amsmath}
\definecolor{hblue}{HTML}{2874A6}
\definecolor{hgreen}{HTML}{28B463}
\definecolor{hpurple}{HTML}{5B2C6F }
\definecolor{horange}{HTML}{DC7633}

\aclfinalcopy

\title{Emotion Stimulus Detection in German News Headlines}

\author{Bao Minh Doan Dang, Laura Oberl\"ander, \and Roman Klinger\\
  Institut f\"ur Maschinelle Sprachverarbeitung, University of
  Stuttgart, Germany\\
  \texttt{st117194@stud.uni-stuttgart.de},\\
  \texttt{\{laura.oberlaender, roman.klinger\}@ims.uni-stuttgart.de} \\
}

\date{}

\begin{document}
\maketitle

\begin{abstract}
Emotion stimulus extraction is a fine-grained subtask of emotion
analysis that focuses on identifying the description of the cause
behind an emotion expression from a text passage (e.g., in the
sentence ``I am happy that I passed my exam'' the phrase ``passed my
exam'' corresponds to the stimulus.). Previous work mainly focused on
Mandarin and English, with no resources or models for German. We fill
this research gap by developing a corpus of 2006 German news headlines
annotated with emotions and 811 instances with annotations of stimulus
phrases. Given that such corpus creation efforts are time-consuming
and expensive, we additionally work on an approach for projecting the
existing English GoodNewsEveryone (\gne) corpus to a
machine-translated German version. We compare the performance of a
conditional random field (CRF) model (trained monolingually on German
and cross-lingually via projection) with a multilingual XLM-RoBERTa
(\xlmr) model. Our results show that training with the German corpus
achieves higher F1 scores than projection. Experiments with \xlmr
outperform their respective \crf counterparts.
\end{abstract}

\section{Introduction}
Emotions are a complex phenomenon that play a central role in our
experiences and daily communications. Understanding them cannot be
accounted by any single area of study since they can be represented
and expressed in different ways, e.g., via facial expressions, voice,
language, or gestures.  In natural language processing, most models
build on top of one out of three approaches to study and understand
emotions, namely basic emotions
\cite{Ekman1992,Strapparava2007,Aman2007}, the valence-arousal model
\cite{Russell1980,Buechel2017} or cognitive appraisal theory
\cite{Scherer2005, Hofmann2020, Hofmann2021}. Emotion classification
in text has received abundant attention in natural language processing
research in the past few years. Hence, many studies have been
conducted to investigate emotions on social media
\citep{Stieglitz2013,Brynielsson2014,Tromp2015}, in literary and
poetry texts \citep{Kim2019,Haider2020} or for analysing song lyrics
\cite{Mihalcea2012,Ferdinan2018,Edmonds2021}. However, previous work
mostly focused on assigning emotions to sentences or text
passages. These approaches do not allow to identify which event,
object, or person caused the emotion (which we refer to as the
\textit{stimulus}).

Emotion stimulus detection is the subtask of emotion analysis which
aims at extracting the stimulus of an expressed emotion. For instance,
in the following example from FrameNet \cite{Fillmore2003} ``Holmes is
happy \ul{having the freedom of the house when we are out}'' one could
assume that \textit{happiness} or \textit{joy} is the emotion in the
text. One could also highlight that the term ``happy'' indicates the
emotion, ``Holmes'' is the experiencer and the phrase ``having the
freedom of the house when we are out'' (underlined) is the stimulus
for the perceived emotion.
Detecting emotion stimuli provides additional information for a better
understanding of the emotion structures (e.g., semantic frames
associated with emotions). More than that, the fact that stimuli are
essential in understanding the emotion evoked in a text is supported
by research in psychology; Appraisal theorists of emotions seem to
agree that emotions include a cognitive evaluative component of an
event \cite{Scherer2005}. Therefore emotion stimulus detection brings
the field of emotion analysis in NLP closer to the state of the art in
psychology.

To the best of our knowledge, there are mostly corpora published for
Mandarin \citep{Lee2010,Gui2014,Gui2016,Gao2017} and English
\citep{Ghazi2015,Mohammad2014,Kim2018,Bostan2020}. We are not aware of
any study that created resources or models for identifying emotion
stimuli in German.  We fill this gap and contribute the \gerstim
(GERman STImulus) corpus with 2006 German news headlines. The
headlines have been annotated for emotion categories, for the mention
of an experiencer or a cue phrase, and for stimuli on the token level
(on which we focus in this paper).  News headlines have been selected
as the domain because they concisely provide concrete information and
are easy to obtain. Additionally, unlike social media texts, this
genre avoids potential privacy issues \cite{Bostan2020}. Given that
annotating such a corpus is time-consuming, we propose a heuristic
method for projecting an annotated dataset from a source language to a
target language. This helps to increase the amount of training data
without manually annotating a huge dataset. Within this study, the
GoodNewsEveryone corpus \citep[\gne,][]{Bostan2020} is selected as an
English counterpart.

Our contributions are therefore: (1) the creation, publication, and
linguistic analysis of the \gerstim dataset to understand the
structure of German stimulus mentions;\footnote{The data is available
  at \url{https://www.ims.uni-stuttgart.de/data/emotion}.} (2), the
evaluation of baseline models using different combinations of feature
sets; and (3) comparison of this in-corpus training with cross-lingual
training via projection and with a pre-trained cross-lingual language
model with XLM-RoBERTa \citep{Conneau2020}.

\section{Related Work}

We now introduce previous work on emotion analysis and for detecting
emotion stimuli.

\subsection{Emotion Analysis}

Emotion analysis is the task of understanding emotions in text,
typically based on psychological theories of \citet{Ekman1992},
\citet{Plutchik2001}, \citet{Russell1980} or
\citet{Scherer2005}. Several corpora have been built for emotion
classification such as \citet{Alm2005} with tales,
\citet{Strapparava2007} with news headlines, \citet{Aman2007} with
blog posts, \citet{Buechel2017} with various domains or \citet{Li2017}
with conversations. Some datasets were created using crowdsourcing, for
instance \citet{Mohammad2014} , \citet{Mohammad2015} or
\citet{Bostan2020}, that have been annotated with tweets, or news
headlines, respectively. Some resources mix various annotation
paradigms, for example \citet{Troiano2019} (self-reporting and
crowd-sourcing) or \citet{Haider2020} (experts and crowdworkers).

Emotion analysis also includes other aspects such as emotion
intensities and emotion roles \citep{Aman2007,Mohammad2017,Bostan2020}
including experiencers, targets, and stimuli \citep{Mohammad2014,
  Kim2018}.

\subsection{Stimulus Detection}

Emotion stimulus detection received substantial attention for Chinese Mandarin
\citep[i.a.]{Lee2010,Li2014,Gui2014,Gui2016,Cheng2017}. Only few corpora have
been created for English \citep{Neviarouskaya2013, Mohammad2014,
Kim2018,Bostan2020}. \citet{Russo2011} worked on a dataset for Italian news
texts and \citet{Yada2017} annotated Japanese sentences from news articles and
question/answer websites.

\citet{Lee2010,Lee2010a} developed linguistic rules to extract emotion
stimuli. A follow-up study developed a machine learning model that
combines different sets of such rules
\citep{Chen2010}. \citet{Gui2014} extended these rules and machine
learning models on their Weibo corpus. \citet{Ghazi2015} formulated
the task as structured learning.

Most methods for stimulus detection have been evaluated on Mandarin.
\citet{Gui2016} propose a convolution kernel-based learning method and
train a classifier to extract emotion stimulus events on the clause
level.  \citet{Gui2017} treat emotion stimulus extraction as a
question answering task.  \citet{Li2018} use a co-attention neural
network. \citet{Chen2018} explore a joint method for emotion
classification and emotion stimulus detection in order to capture
mutual benefits across these two tasks. Similarly, \citet{Xia2019}
evaluate a hierarchical recurrent neural network transformer model to
classify multiple clauses. They show that solving these subtasks
jointly is beneficial for the model's performance.

\citet{Xia2019a} redefine the task as emotion/cause pair extraction
and intend to detect potential emotions and corresponding causes in
text. \citet{Xu2019} tackle the emotion/cause pair extraction task by
adopting a learning-to-rank method. \citet{Wei2020} also argue for the
use of a ranking approach. They rank each possible emotion/cause pair
instead of solely ranking stimulus phrases. \citet{Fan2020} do not
subdivide the emotion/cause pair detection task into two subtasks but
propose a framework to detect emotions and their associated causes
simultaneously.

\citet{Oberlaender2020} studied whether sequence labeling or clause
classification is appropriate for extracting English stimuli. As we assume that
these findings also hold for German, we follow their finding that token
sequence labeling is more appropriate.

\section{Corpus Creation}

To tackle German emotion stimulus detection on the token-level, we select
headlines from various online news portals, remove duplicates and irrelevant
items, and further subselect relevant instances with an emotion dictionary. Two
annotators then label the data. We describe this process in detail in the
following.

\subsection{Data Collection}

We select various German news sources and their RSS feeds based on
listings at a news overview
website\footnote{\url{https://www.deutschland.de/de/topic/wissen/nachrichten},
  accessed on April 27, 2021} and add some regional online
newspapers.\footnote{The list of RSS feeds is available in the
  supplemental material.}  The collected corpus consists of headlines
between September 30, 2020 and October 7, 2020 and between October
22 and October 23, 2020 with 9000 headlines, spread across several domains
including \emph{politics}, \emph{sports}, \emph{tech and business},
\emph{science} and \emph{travel}.

\subsection{Data Preprocessing and Filtering}
Short headlines, for instance ``Verlobung!'' or ``Krasser After-Baby-Body'' do
not contain sufficient information for our annotation, therefore we omit
sentences that have less than 5 words. Further, we remove generic parts of the
headline, like ``++ Transferticker ++'', ``+++ LIVE +++'' or ``News-'' and only
keep the actual headline texts.

We also remove headlines that start with particular key words which denote a
specific event which would not contribute to an understanding of emotions or
stimuli, such as ``Interview'', ``Kommentare'', ``Liveblog'', ``Exklusive'', as
well as visual content like ``Video'', ``TV'' or ``Pop''. Additionally, we discard
instances which include dates, like ``Lotto am Mittwoch, 30.09.2020'' or
``Corona-News am 05.10''.\footnote{Details in Supplementary Material.}

After filtering, we select instances that are likely to be associated with an
emotion with the help of an emotion lexicon \citep{Klinger2016}. For this
purpose, we accept headlines which include at least one entry from the
dictionary.

\subsection{Annotation}
The annotation of the 2006 headlines which remain after preprocessing
and filtering consists of two phases. In the first phase, emotion
cues, experiencers and emotion classes are annotated, while stimuli
are addressed in the second phase only for those instances which
received an emotion label. Table~\ref{tb:questionaires} in the
Appendix shows the questions to be answered during this annotation
procedure. Each headline in the dataset is judged by two
annotators. One of them is female (23 years old) while the other
annotator is male (26 years old).  The first annotator has a
background in digital humanities and linguistics, while the second has
a background in library and information management. After each phase,
we combine overlapping stimulus annotations by choosing the parts
annotated by both annotators, and discuss the cases where the
annotations do not overlap until a consensus is reached.

\begin{table}[t]
  \centering\small
  \begin{tabularx}{\linewidth}{lX}
    \toprule
    No. & Linguistics Rules \\
    \cmidrule(l{0pt}r{5pt}){1-1} \cmidrule(l{5pt}){2-2}
    1. & Stimuli can be described by verbal  or nominal phrases\\
    2. & Subjunctions like ``because of'' belong to the sequence\\
    3. & Conjunctions like ``and'', ``or'' and ``but'' connect main clauses. They can therefore belong to a stimulus sequence.\\
    4. & Antecedents, if present, are annotated as stimuli\\
    5. & If antecedent is not present, an anaphora may be annotated instead\\
    6. & Composites with ``-'' are considered a single word\\
    7. & Stimuli can include one or multiple words\\
    8. & Punctuation (e.g. ,.-:;``''!?) should not be labeled as stimulus\\
    \bottomrule
  \end{tabularx}
  \caption{Linguistics rules for annotating stimuli.}
  \label{tb:ling_rules}
\end{table}

\begin{table}
  \centering\small
  \begin{tabular}{lrrrrrr}
    \toprule
    &&&&& \multicolumn{2}{c}{\F}\\
    \cmidrule(l){6-7}
    & \multicolumn{4}{c}{$\kappa$} & tok. & span  \\
    \cmidrule(r){2-5}\cmidrule(lr){6-6}\cmidrule(l){7-7}
    Iteration & Cue & Exp. & Emo. & \multicolumn{3}{c}{Stim.} \\
    \cmidrule(r){1-1}\cmidrule(lr{5pt}){2-2}\cmidrule(lr){3-3}\cmidrule(lr){4-4}\cmidrule(lr ){5-7}
    Prelim. 1 & .22 & .43  & .25  & --- & --- & ---\\
    Prelim. 2 & .71 & .49  & .47  & --- & --- & ---\\
    Prelim. 3 & .46 &  .69 & .44  & --- & .65 & --- \\
    \cmidrule(r){1-1}\cmidrule(lr{5pt}){2-2}\cmidrule(lr){3-3}\cmidrule(lr){4-4}\cmidrule(lr){5-7}
    Final     & .56 & .57  & .51  & .68 & .72 & .56 \\
    \bottomrule
  \end{tabular}
  \caption{Inter-annotator agreement for the binary tasks of annotating
    the existance of cue mentions, experiencer mentions, the
    multi-label annotation of emotion labels, and the token-level
    annotation of stimulus spans. The \F-span value for stimuli is an
    exact match value for the whole span. }
  \label{tb:sample_anno}
\end{table}

\paragraph{Guidelines.} We created an initial version of guidelines
motivated by \newcite{Lee2010,Lee2010a,Gui2014,Ghazi2015}. Based on
two batches of 25 headlines, and one with 50 headlines, we refined the
guidelines in three iterations. After each iteration, we calculated
inter-annotator agreement scores and discussed the annotator's
results. It should be noted that we only considered annotating
emotions in the first two iterations.  The sample annotation of
emotion stimuli on the token-level has been performed in the third
round, i.e., after two discussions and guideline refinements.  During
these discussions, we improved the formulation of the annotation task,
provided more detailed descriptions for each predefined emotion and
clarified the concept of sequence labeling using the IOB
scheme. Additionally, we formulated several linguistic rules that help
annotating stimuli (see Table \ref{tb:ling_rules}).

\paragraph{Details.}
The goal of Phase 1 of the annotation procedure is to identify
headlines with an emotional connotation. Those which do then receive
stimulus annotations in Phase 2.

We annotated in a spread sheet application.  In Phase 1a both
annotators received 2006 headlines. They were instructed to annotate
whether a headline expresses an emotion by judging if cue words or
experiencers are mentioned in the text. Further, only one, the most
dominant, emotion is to be annotated (\emph{happiness},
\emph{sadness}, \emph{fear}, \emph{disgust}, \emph{anger},
\emph{positive surprise}, \emph{negative surprise}, \emph{shame},
\emph{hope}, \emph{other} and \emph{no emotion}).  In Phase 1b we
aggregated emotion annotations and jointly discussed non-overlapping
labels to a consensus annotation.

In Phase 2a, annotators were instructed to label pretokenized
headlines with the IOB alphabet for stimulus spans -- namely those
which received an emotion label in Phase 1 (811 instances). In Phase
2b, we aggregated the stimulus span annotations to a gold standard by
accepting all overlapping tokens of both annotators in cases where
they partially matched. For the other cases where the stimulus
annotations did not overlap, we discussed the annotations to reach an
agreement.

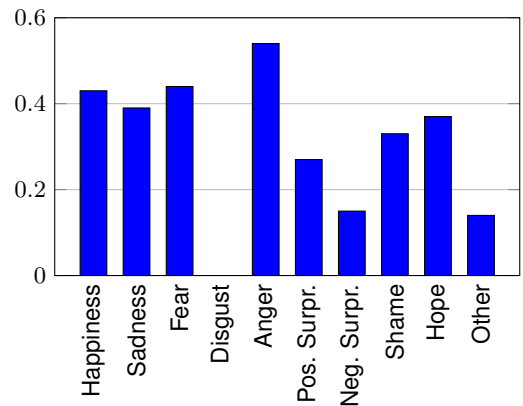
\begin{figure}[t]
  \centering\small\sffamily
  \begin{tikzpicture}
    \begin{axis}[
      symbolic x coords={Happiness, Sadness, Fear, Disgust, Anger, Pos. Surpr., Neg. Surpr., Shame, Hope, Other},
      ymin=0,
      ymax=0.6,
      xticklabel style = {rotate=90}, xtick=data,
      ymajorgrids = true,
      major x tick style = transparent,
      width=\linewidth,
      height=5cm
      ]
      \addplot[ybar,fill=blue] coordinates {
        (Happiness, 0.43)
        (Sadness, 0.39)
        (Fear, 0.44)
        (Disgust,0. 0)
        (Anger, 0.54)
        (Pos. Surpr.,0. 27)
        (Neg. Surpr., 0.15)
        (Shame, 0.33)
        (Hope, 0.37)
        (Other,0.14)
      };
    \end{axis}
  \end{tikzpicture}
  \caption{$\kappa$ for all emotion classes.}
  \label{fg:emotion_agree}
\end{figure}

\paragraph{Agreement Results.} Table \ref{tb:sample_anno} presents the
inter-annotator agreement scores for the preliminary annotation rounds and for
the final corpus. We observe that the results are moderate across classes.
Figure \ref{fg:emotion_agree} illustrates the agreement for each emotion class.
The emotions \emph{anger}, \emph{fear}, and \emph{happiness} show the highest
agreement, while \emph{surprise}, \emph{other}, and particularly \emph{disgust}
show lower scores.

For the stimulus annotation, we evaluate the agreement via token-level
Cohen's $\kappa$, via token-level \F, and via exact span-match \F (in
the first two cases, B and I labels are considered to be different).
The token-level result for the final corpus is substantial with
$\kappa$ =.68, \F=.72 and moderate for the exact span match, with
\F=.56 (see Table~\ref{tb:sample_anno}).

\section{Corpus Analysis}
\begin{table}[t]
  \centering
  \scalebox{0.95}{\begin{tabular}{lrrrrrr}
    \toprule
    Emotion         & \rt{\# \ inst.} & \rt{w/ cue} & \rt{w/ exp} & \rt{w/ stimulus} & \rt{avg.\ $|$stimulus$|$} \\
    \cmidrule(r){1-1}\cmidrule(rl){2-2}\cmidrule(rl){3-3}\cmidrule(rl){4-4}\cmidrule(rl){5-5}\cmidrule(l){6-6}
    Happiness       &    80   &  80    & 77   & 76  & 3.72\\
    Sadness         &    65   &  65    & 54   & 59  & 4.07\\
    Fear            &   177   &   117  & 138  & 167 & 3.83 \\
    Disgust         &      3  &   3    & 2    & 3   & 4.00\\
    Anger           &     226 &  226   & 195  & 208 & 3.86 \\
    Pos.\ Surprise  &    51   &   51   &  45  & 44  & 4.11\\
    Neg.\ Surprise  &   142   &  140   & 125  & 130 & 3.96\\
    Shame           &     9   &   9    & 9    & 8   & 3.75\\
    Hope            &    20   &  19    & 16   & 19  & 4.05\\
    Other           &    38   &   37   & 26   & 34  & 3.71\\
    No Emo.         &  1195   &  930   & 109  &  -   & - \\
    \cmidrule(r){1-1}\cmidrule(rl){2-2}\cmidrule(rl){3-3}\cmidrule(rl){4-4}\cmidrule(rl){5-5}\cmidrule(lr){6-6}
    All             &  2006 & 1737 & 796 &  748& 3.9&  \\
    \bottomrule
  \end{tabular}
    }
  \caption{Corpus statistics. Columns show the amount of annotated instances for emotion cue,
  	  experiencer, stimulus and the average length of all stimulus spans within each respective
  	  dominant emotion. For aggregating cue and experiencer, cases where one of the annotators
  	 annotated with a \textit{yes} have been accepted. }
  \label{tab:corpusstat}
\end{table}

\begin{table}[t]
  \centering\small
  \begin{tabularx}{\linewidth}{lX}
    \toprule
    Emotion & News Sources\\
    \cmidrule(r){1-1}\cmidrule(l){2-2}
    Happiness & Bild, Welt, Stuttgarter Zeitung\\
    Sadness & Bild, Spiegel, Stuttgarter Z. \\
    Fear & Stuttgarter Z., Bild, Welt\\
    Disgust & T-Online, Welt, Spiegel\\
    Anger & Bild, Stuttgarter Z., Spiegel\\
    Pos. Surprise & Welt, Focus, Bild \\
    Neg. Surprise & Bild, Stuttgarter Z., Spiegel\\
    Shame & Stuttgarter Z., Bild,  Welt\\
    Hope & T-Online, Bild, Stuttgarter Z.\\
    Other & Bild, Stuttgarter Z., Welt\\
    \bottomrule
  \end{tabularx}
  \caption{Top three most observed media sources for each dominant emotion sorted by frequency.}
  \label{tb:top_emotion_source}
\end{table}

\subsection{Quantitative Analysis}
Our corpus consists of 2006 headlines with 20,544 tokens and 6,763
unique terms.  From those, 811 instances were labeled with an emotion
category and received stimulus annotations on the token-level. The
shortest headline consists of five words, while the longest has 20
words. The headlines are on average short with nine words. The
stimulus spans range from one to eleven tokens and have four words on
average.

Table~\ref{tab:corpusstat} summarizes the corpus statistics of \gerstim. For
aggregating emotion cue and experiencer we accept instances for which the
mention of these emotion roles has been annotated by one annotator. For all
emotions, most instances include the mention of an emotion cue (likely biased
by our sampling procedure). Further, the number of headlines with mentions of a
stimulus and an experiencer is also high for those instances which are labeled
to be associated with an emotion.

Table~\ref{tb:top_emotion_source} presents the most common sources,
sorted by their frequencies, for each aggregated emotion during Phase
1b. Not surprisingly, \textit{Bild-Zeitung} is to be found in the top
three for almost all emotion classes, followed by
\textit{Stuttgarter-Zeitung} and \textit{Welt}. In particular, in five
out of ten of the emotions, \textit{Bild-Zeitung} takes the first
place. As Table~\ref{tab:corpusstat} demonstrates, \textit{disgust} is
relatively rare, we therefore list all available sources for this
emotion category. Furthermore, four in five most frequently annotated
emotions are negative (\textit{anger}, \textit{fear}, \textit{negative
  surprise}, \textit{happiness}, \textit{sadness}).

Note that this analysis does not necessarily reflect the actual quality
of chosen news sources. The findings we report here might strongly be
biased by the data collection time span.

\subsection{Qualitative Analysis of Stimuli}
\label{sec:qual}

To obtain a better understanding of stimuli in German, we analyse which words
together with their preferred grammatical realizations are likely to indicate a
stimulus phrase. For this purpose, we examine the parts of speech\footnote{We
    use spaCy, \url{https://spacy.io/usage/linguistic-features}, accessed on
April 29, 2021} of terms that are directly left positioned to stimulus phrases, inside
the stimulus phrases and right after it (see Table \ref{tb:pos}).
We further compare our findings with Mandarin \cite{Lee2010a} and English
\cite{Bostan2020}.

Our analysis shows that for \gerstim common nouns, proper nouns, punctuation, and verbs are
most frequently located directly to the left of stimulus mentions (common nouns
$\approx$26\%, punctuation $\approx$28\%, verbs $\approx$22\%, proper nouns $\approx$0.09\%). Often, these
words are emotionally connotated, for instance as in the nouns ``Streit'',
``Angst'', ``Hoffnung'' or ``Kritik'' or the verbs ``warnen'', ``kritisieren'',
``bedrohen'', ``beklagen'' or ``k\"ampfen''.

\begin{table*}[t]
  \centering\small
  \setlength{\tabcolsep}{8pt}
\begin{tabular}{lcccccccc}
\toprule
& \multicolumn{4}{c}{GERSTI} & \multicolumn{4}{c}{GNE} \\
\cmidrule(rl){2-5}\cmidrule(l){6-9}
    POS & All & Inside & Before@1 & After@1  & All & Inside & Before@1 & After@1  \\
\cmidrule(r){1-1}\cmidrule(rl){2-2}\cmidrule(rl){3-3}\cmidrule(rl){4-4}\cmidrule(rl){5-5}\cmidrule(rl){6-6}\cmidrule(rl){7-7}\cmidrule(rl){8-8}\cmidrule(l){9-9}
    \texttt{NOUN}  & .28 & .33 (1.17$\times$) & .26 (0.93$\times$) & .00 (0.01$\times$) & .16 & .17 (1.09$\times$) & .11 (0.69$\times$) & .17 (1.05$\times$) \\
    \texttt{ADP}   & .15 & .22 (1.48$\times$) & .03 (0.19$\times$) & .23 (1.54$\times$) & .10 & .12 (1.12$\times$) & .14 (1.37$\times$) & .20 (1.95$\times$) \\
    \texttt{PROPN} & .14 & .09 (0.65$\times$) & .09 (0.68$\times$) & .01 (0.04$\times$) & .30 & .26 (0.89$\times$) & .25 (0.86$\times$) & .25 (0.83$\times$) \\
    \texttt{PUNCT} & .13 & .02 (0.16$\times$) & .28 (2.23$\times$) & .49 (3.87$\times$) & .09 & .07 (0.82$\times$) & .21 (2.40$\times$) & .08 (0.91$\times$) \\
    \texttt{VERB}  & .09 & .09 (0.91$\times$) & .22 (2.32$\times$) & .16 (1.68$\times$) & .11 & .12 (1.06$\times$) & .09 (0.80$\times$) & .09 (0.85$\times$) \\
    \texttt{DET}   & .05 & .08 (1.47$\times$) & .00 (0.09$\times$) & .01 (0.16$\times$) & .04 & .05 (1.03$\times$) & .04 (0.81$\times$) & .03 (0.63$\times$) \\
    \texttt{ADJ}   & .05 & .07 (1.44$\times$) & .00 (0.03$\times$) & .01 (0.29$\times$) & .05 & .05 (1.09$\times$) & .02 (0.42$\times$) & .03 (0.53$\times$) \\
    \texttt{ADV}   & .05 & .05 (1.04$\times$) & .04 (0.87$\times$) & .04 (0.93$\times$) & .02 & .02 (1.07$\times$) & .02 (0.80$\times$) & .03 (1.47$\times$) \\
    \texttt{AUX}   & .02 & .01 (0.75$\times$) & .04 (2.34$\times$) & .03 (1.68$\times$) & .03 & .03 (1.01$\times$) & .03 (1.16$\times$) & .03 (1.11$\times$) \\
    \texttt{PRON}  & .01 & .01 (0.71$\times$) & .02 (1.02$\times$) & .00 (0.19$\times$) & .03 & .03 (1.14$\times$) & .01 (0.45$\times$) & .02 (0.63$\times$) \\
    \texttt{NUM}   & .01 & .02 (1.49$\times$) & .00 (0.00$\times$) & .00 (0.00$\times$) & .02 & .02 (1.15$\times$) & .01 (0.27$\times$) & .01 (0.34$\times$) \\
    \texttt{CCONJ} & .01 & .01 (0.97$\times$) & .01 (0.55$\times$) & .01 (0.77$\times$) & .01 & .01 (1.21$\times$) & .00 (0.64$\times$) & .02 (3.82$\times$) \\
\bottomrule
\end{tabular}
\caption{Relative frequencies of POS tags of all tokens in GERSTI and
  GNE datasets (All) vs relative frequencies of POS tags inside the
  stimuli spans (Inside), before and after the stimuli spans
  (Before@1, After@1). For all the columns that show frequencies of
  the spans related to the stimuli we show the factor ($\times$) of
  how much it differs to the global frequencies in All.}
\label{tb:pos}
\end{table*}

There are discrepancies between German and Mandarin
stimuli. \citet{Lee2010a,Lee2010} state that prepositions or
conjunctions mostly indicate stimulus phrases in Mandarin, while this
is not the case for German due to our predefined annotation rules
(Rule 2 from Table \ref{tb:ling_rules}). Furthermore, indicator words
for Chinese stimulus events do not cover common nouns or proper
nouns. However, verbs seem to emphasize emotion causes in both
languages.

Compared to \gne, we also notice some differences: English stimuli do not begin
with prepositions, but prepositions are most likely to be included in the
stimulus span (\texttt(ADP) $\approx$0.14\% in GNE vs $\approx$0.03\% in
\gerstim). Further, by looking at the part of speech tags that were relevant in
indicating the stimuli for \gerstim we see that they are dominating for GNE
as well. However, there are far more proper nouns than common nouns and quite
fewer verbs that occur right before the stimulus phrase (common nouns$\approx$11\%,
punctuation $\approx$21\%, verbs $\approx$0.09\%, proper nouns
$\approx$0.25\%). Often, these indicator words of English stimuli do not as
directly evoke an emotion. For instance, ``say'', ``make'', ``woman'',
``people'' or ``police'' are often observed to be directly left located words
of English stimuli. Nevertheless, similar to \gerstim, stimuli from \gne corpus
are not indicated by conjunctions, numerals or pronouns.

The positioning of the stimuli is only similar to a limited degree in German and
English: 53\% of the instances in \gerstim end with the
stimulus (86\% in English \gne) and 13\% begin with the
stimulus (11\% in \gne).

\section{Experiments}
In the following, we explain how we project annotation from an English
stimulus corpus to a machine-translated counterpart. Based on this, we
evaluate how well a linear-chain conditional random field
\cite{Lafferty2001} performs with the projected dataset
in comparison to the monolingual setup.
We compare that result to the use of the pre-trained
language model \textit{XLM-RoBERTa} (\xlmr) \citep{Conneau2020}.

\subsection{Annotation Projection}

We use the  \gne dataset \citep{Bostan2020} which is a large English annotated
corpus of news headlines. Stimulus sequences  in this dataset are comparatively
longer with eight tokens on average.

We translate the GNE corpus via
\textit{DeepL}\footnote{\url{https://www.deepl.com/en/translator}, accessed on
May 20, 2021} and perform the annotation projection as follows: We first
translate the whole source instance $\mathbf{t}_{\mathrm{en}}$ to the
translation $\mathbf{t}_{\mathrm{de}}$ (from English to German). We further
translate the stimulus token sequence $\mathbf{stim}_{\mathrm{en}}$ to
$\mathbf{stim}_{\mathrm{de}}$. We assume the stimulus annotation for
$\mathbf{t}_{\mathrm{de}}$ to correspond to all tokens in
$\mathbf{stim}_{\mathrm{de}}$, heuristically corrected to be a consecutive
sequence.

\subsection{Experimental Setting}
\subsubsection{Models}
\paragraph{CRF.} We implement the linear-chain conditional random
field model via the CRF-suite in
Scikit-learn\footnote{\url{https://sklearn-crfsuite.readthedocs.io/en/latest/},
 accessed on April 30, 2021} and extract different features. What we
call \emph{corpus-based features} contains the frequency of a current
word in the whole corpus, position label for first (\emph{begin}),
last (\emph{end}) and remaining (\emph{middle}) words of the
headline, if the current word is capitalized, or entirely in
upper or lowercase, if the token is a number, a punctuation symbol, or in
the list of 50 most frequent words in our corpus.

We further include \emph{linguistic features}, namely the
part-of-speech tag, the syntactic dependency between the current token
and its head, if it is a stopword or if it has a named entity label
(and which one it is).

We further add a feature which specifies whether the token is part of
an emotion-word dictionary \citep{Klinger2016}. Additionally, we
combine the feature vector of the preceding and succeeding token (we
add the prefixes \emph{prev} and \emph{next} to each feature name)
with the current token to get information about surrounding words. We
mark the first and last token with additional features.

\paragraph{RoBERTa.} We use the pre-trained \textit{XLM-RoBERTa} base
model with the
\textit{HuggingFace}\footnote{\url{https://huggingface.co/xlm-roberta-base},
  accessed on April 30, 2021} library from \citet{Wolf2020}. In
addition to the pre-trained transformer, we add a linear layer which
outputs a sequence of IOB tags for each input sentence. We fine-tune
the language model in five epochs and use a batch size of 16 during
training, a dropout rate of 0.5, and the Adam optimizer with weight
decay \cite{Loshchilov2017}, with a learning rate of $10^{-5}$ and a
maximum gradient norm of 1.0.

\paragraph{Setup.}
For our experiments, we only use the 811 instances from the \gerstim
dataset that received annotations for emotion stimuli.  We split them
into a train and validation subset (80\,\%/20\,\%) and perform
experiments in three different settings. In the \emph{in-corpus
  training}, we train with the \gerstim training data and test on the
test corpus. In the \emph{projection} setting, we train on the english
\gne data and test on the German \gerstim test data (either with the
CRF via projection or directly with the \xlmr model). In the
\emph{aggregation} setting, we use both the English train data and the
German train data for training.

\subsubsection{Evaluation Metrics}

We evaluate the stimuli prediction as follows (following \citet{Ghazi2015} and
\citet{Oberlaender2020}): \emph{Exact} match leads to a true positive for an
exactly correct span prediction. \emph{Partial} accepts a predicted stimulus as
true positive if at least one token overlaps with a gold standard span. A
variation is \emph{Left}/\emph{Right}, where the left/right boundary needs to
perfectly match the gold standard.

\subsection{Results}

Table \ref{tb:crfs} reports the results for our experiments. The top
four blocks compare the importance of the feature set choice for the
CRF approach.

In nearly all combinations of model and evaluation measure, the
in-corpus evaluation leads to the best performance -- adding data from
the \gne corpus only slightly improves for the \emph{Partially} evaluation
setting when the CRF is limited to corpus features. The
projection-based approach, where the model does not have access to the
\gerstim training data consistently shows a lower performance, with
approximately a drop by 50\,\% in \F score.

The linguistic features particularly help the CRF in the \emph{Exact}
evaluation setting, but all feature set choices are dominated by the
results of the XLM-RoBERTa model. This deep learning approach shows the
best results across all models, and is particularly better in the
\emph{Partial} evaluation setting, with 19pp, 13pp and 15pp
improvement.

Both projection and aggregation models indicate that extracting the beginning
of a stimulus span is challenging. We assume that both models have learned
English stimulus structures and therefore could not generalize well on the
German emotion stimuli (also see Section~\ref{sec:qual}).

\begin{table}[t]
  \centering\small
  \begin{tabularx}{\linewidth}{Xlccc}
    \toprule
    Model & $F_1$ & in-corp. & proj. & aggre. \\
    \cmidrule(r){1-1}\cmidrule(lr){2-2}\cmidrule(lr){3-3}\cmidrule(lr){4-4}\cmidrule(l){5-5}
    \multirow{4}{*}{\parbox{15mm}{CRF with corpus features}}
    & Exact    & \textbf{.38} & .19 & .33\\
    & Partial  & .49 & .43 & \textbf{.52}\\
    & Left     & \textbf{.42} & .22 & .38\\
    & Right    & \textbf{.51} & .41 & \textbf{.51}\\
    \cmidrule(r){1-1}\cmidrule(lr){2-2}\cmidrule(lr){3-3}\cmidrule(lr){4-4}\cmidrule(l){5-5}
    \multirow{4}{*}{\parbox{15mm}{CRF with linguistic features}}
    & Exact    & \textbf{.42} & .16 & .35\\
    & Partial  & \textbf{\textit{.58}} & .41 & .54\\
    & Left     & \textbf{.52} & .19 & \textit{.43}\\
    & Right    & \textbf{\textit{.57}} & .40 & \textit{.53}\\
    \cmidrule(r){1-1}\cmidrule(lr){2-2}\cmidrule(lr){3-3}\cmidrule(lr){4-4}\cmidrule(l){5-5}
    \multirow{4}{*}{\parbox{15mm}{CRF with corp.+lingu. features}}
    & Exact    & \textbf{\textit{.45}} & .19 & .35\\
    & Partial  & \textbf{.57} & \textit{.48} & .53\\
    & Left     & \textbf{\textit{.53}} & .24 & .41\\
    & Right    & \textbf{.56} & \textit{.47} & .52\\
    \cmidrule(r){1-1}\cmidrule(lr){2-2}\cmidrule(lr){3-3}\cmidrule(lr){4-4}\cmidrule(l){5-5}
    \multirow{4}{*}{\parbox{15mm}{CRF with all features}}
    & Exact    & \textbf{.42} & \textit{.20} & \textit{.36}\\
    & Partial  & \textbf{.56} & \textit{.48} & \textit{.55}\\
    & Left     & \textbf{.50} & \textit{.25} & \textit{.43}\\
    & Right    & \textbf{.55} & .46 & \textit{.53}\\
    \cmidrule(r){1-1}\cmidrule(lr){2-2}\cmidrule(lr){3-3}\cmidrule(lr){4-4}\cmidrule(l){5-5}
    \multirow{4}{*}{\parbox{15mm}{RoBERTa XLM-R}}
    & Exact    & \textbf{\underline{.47}}& \underline{.25}& \underline{.45} \\
    & Partial  & \textbf{\underline{.75}}& \underline{.61}& \underline{.70} \\
    & Left     & \textbf{\underline{.68}}& \underline{.35}& \underline{.58} \\
    & Right    & \underline{.71}& \underline{.59}& \textbf{\underline{.72}} \\
    \bottomrule
  \end{tabularx}
  \caption{Results for the CRF models with different feature sets and
    the \xlmr model. Highest \F-scores in each row printed
    with \textbf{bold face}, highest score in column/per evaluation
    measure is \underline{underlined}, highest score in each column
    and per evaluation measure in the CRF is printed
    \textit{italics}.}
  \label{tb:crfs}
\end{table}

\begin{table*}[t]
	\centering
	\begin{tabularx}{\textwidth}{lXl}
		\toprule
		Err. Type & Example & Setup \\
		\cmidrule(r){1-1} \cmidrule(lr){2-2}\cmidrule(lr){3-3}
		Early start &  $\underset{\textrm{Court}}{\hbox{Hof}\vphantom{y}}$ $\underset{\textrm{in}}{\hbox{in}\vphantom{y}}$ $\underset{\textrm{Bavaria}}{\hbox{Bayern}\vphantom{y}}$$\underset{\textrm{:}}{\hbox{:}\vphantom{y}}$ \textcolor{red}{$\underset{\textrm{21-year-old}}{\hbox{21-J\"ahriger}\vphantom{y}}$} \leftbr \textcolor{blue}{$\underset{\textrm{after}}{\hbox{nach}\vphantom{y}}$ $\underset{\textrm{deadly}}{\hbox{t\"odlichem}\vphantom{y}}$ $\underset{\textrm{car-accident}}{\hbox{Autounfall}\vphantom{y}}$ $\underset{\textrm{to}}{\hbox{zu}\vphantom{y}}$ $\underset{\textrm{probation}}{\hbox{Bew\"ahrungsstrafe }\vphantom{y}}$ $\underset{\textrm{convicted}}{\hbox{verurteilt }\vphantom{y}}$} \rightbr  &  projection\\
		& \textit{Court in Bavaria: 21-year-old sentenced to probation after fatal car accident} & \\
		\cmidrule(r){1-1} \cmidrule(lr){2-2}\cmidrule(lr){3-3}
		
		Late start & $\underset{\textrm{Peter}}{\hbox{Peter}\vphantom{y}}$ $\underset{\textrm{Madsen}}{\hbox{Madsen}\vphantom{y}}$ $\underset{\textrm{in}}{\hbox{in}\vphantom{y}}$ $\underset{\textrm{Denmark}}{\hbox{D\"anemark}\vphantom{y}}$$\underset{\textrm{:}}{\hbox{:}\vphantom{y}}$ $\underset{\textrm{Kim}}{\hbox{Kim}\vphantom{y}}$ $\underset{\textrm{Wall's}}{\hbox{Walls}\vphantom{y}}$ $\underset{\textrm{murderer}}{\hbox{Mörder}\vphantom{y}}$
		\leftbr  $\underset{\textrm{fails}}{\hbox{scheitert}\vphantom{y}}$ \textcolor{red}{$\underset{\textrm{by}}{\hbox{bei}\vphantom{y}}$} \textcolor{blue}{$\underset{\textrm{escape-attempt}}{\hbox{Fluchtversuch}\vphantom{y}}$ $\underset{\textrm{from}}{\hbox{aus}\vphantom{y}}$ $\underset{\textrm{prison}}{\hbox{Gefängnis}\vphantom{y}}$} \rightbr  &in-corpus \\
		& \textit{Peter Madsen from Denmark: Kim Wall's killer fails in escape attempt from prison}& \\
		\cmidrule(r){1-1} \cmidrule(lr){2-2}\cmidrule(lr){3-3}
		
		Early stop &  $\underset{\textrm{Even}}{\hbox{Noch}\vphantom{y}}$ $\underset{\textrm{more}}{\hbox{mehr}\vphantom{y}}$ $\underset{\textrm{parents}}{\hbox{Eltern}\vphantom{y}}$ $\underset{\textrm{tell}}{\hbox{erzählen}\vphantom{y}}$
		\leftbr  \textcolor{red}{$\underset{\textrm{about}}{\hbox{von}\vphantom{y}}$} \textcolor{blue}{$\underset{\textrm{the}}{\hbox{den}\vphantom{y}}$ $\underset{\textrm{scary}}{\hbox{unheimlichen}\vphantom{y}}$ $\underset{\textrm{things}}{\hbox{Dingen}\vphantom{y}}$}$\underset{\textrm{,}}{\hbox{,}\vphantom{y}}$ $\underset{\textrm{that}}{\hbox{die}\vphantom{y}}$ $\underset{\textrm{their}}{\hbox{ihr}\vphantom{y}}$ $\underset{\textrm{child}}{\hbox{Kind}\vphantom{y}}$ $\underset{\textrm{once}}{\hbox{mal}\vphantom{y}}$ $\underset{\textrm{said}}{\hbox{gesagt}\vphantom{y}}$ $\underset{\textrm{has}}{\hbox{hat}\vphantom{y}}$ \rightbr & in-corpus\\
		& \textit{More parents share creepy things their kid once said} & \\
		\cmidrule(r){1-1} \cmidrule(lr){2-2}\cmidrule(lr){3-3}
		
		Late stop  &  $\underset{\textrm{In}}{\hbox{In}\vphantom{y}}$ $\underset{\textrm{Paris}}{\hbox{Paris}\vphantom{y}}$$\underset{\textrm{:}}{\hbox{:}\vphantom{y}}$ \leftbr\textcolor{red}{$\underset{\textrm{Loud}}{\hbox{Lauter}\vphantom{y}}$} \textcolor{blue}{$\underset{\textrm{bang}}{\hbox{Knall}\vphantom{y}}$} \rightbr \textcolor{blue}{$\underset{\textrm{scares}}{\hbox{schreckt}\vphantom{y}}$ $\underset{\textrm{people}}{\hbox{Menschen}\vphantom{y}}$ $\underset{\textrm{on}}{\hbox{auf}\vphantom{y}}$} $\underset{\textrm{-}}{\hbox{-}\vphantom{y}}$ $\underset{\textrm{Cause}}{\hbox{Ursache}\vphantom{y}}$ $\underset{\textrm{quickly}}{\hbox{schnell}\vphantom{y}}$ $\underset{\textrm{found}}{\hbox{gefunden}\vphantom{y}}$& aggregation\\
		& \textit{In Paris: Loud bang startles people - cause quickly found}&\\
		
		\cmidrule(r){1-1} \cmidrule(lr){2-2}\cmidrule(lr){3-3}
		Surrounding  &$\underset{\textrm{EU-summit}}{\hbox{EU-Gipfel}\vphantom{y}}$$\underset{\textrm{:}}{\hbox{:}\vphantom{y}}$ \textcolor{red}{$\underset{\textrm{Dispute}}{\hbox{Streit}\vphantom{y}}$}\leftbr \textcolor{blue}{$\underset{\textrm{about}}{\hbox{\"uber}\vphantom{y}}$ $\underset{\textrm{line}}{\hbox{Linie}\vphantom{y}}$ $\underset{\textrm{to}}{\hbox{zur}\vphantom{y}}$ $\underset{\textrm{Turkey}}{\hbox{T\"ukei}\vphantom{y}}$} \rightbr \textcolor{blue}{$\underset{\textrm{-}}{\hbox{-}\vphantom{y}}$ $\underset{\textrm{Erdogan}}{\hbox{Erdogan}\vphantom{y}}$ $\underset{\textrm{reacts}}{\hbox{reagiert}\vphantom{y}}$ $\underset{\textrm{with}}{\hbox{mit}\vphantom{y}}$ $\underset{\textrm{gloat}}{\hbox{H\"ame}\vphantom{y}}$}& projection\\
		& \textit{EU-summit: Dispute over line on Turkey - Erdogan responds with gloating} & \\
		\cmidrule(r){1-1} \cmidrule(lr){2-2}\cmidrule(lr){3-3}
		
		Consecutive & $\underset{\textrm{Defeat}}{\hbox{Niederlage}\vphantom{y}}$ $\underset{\textrm{for}}{\hbox{f\"ur}\vphantom{y}}$ $\underset{\textrm{car-manufacturer}}{\hbox{Autohersteller}\vphantom{y}}$$\underset{\textrm{:}}{\hbox{:}\vphantom{y}}$ \leftbr\textcolor{red}{$\underset{\textrm{work-council-election}}{\hbox{Betriebsratswahl}\vphantom{y}}$} $\underset{\textrm{by}}{\hbox{bei}\vphantom{y}}$ $\underset{\textrm{Daimler}}{\hbox{Daimler}\vphantom{y}}$ $\;\;\;$ \textcolor{blue}{$\underset{\textrm{invalid}}{\hbox{ung\"ultig}\vphantom{y}}$} \rightbr  & aggregation \\
		& \textit{Defeat for car manufacturer: Daimler's work council election invalid}& \\
		\bottomrule
	\end{tabularx}
	\caption{Example headlines for examined error types. Gold
		annotations correspond to tokens between $\left[\;\right]$. Predicted
		stimulus segments are highlighted as follows: red (B tag), blue (I
		tag). English translations for each sample are written in \textit{italics}. All examples stem from the CRF models except the last
		one.}
	\label{tb:errors_exp}
\end{table*}

\subsection{Error Analysis}

We now discuss the model's quality (see Table \ref{tb:errors_exp})
based on various error types, namely \emph{Early Start}, \emph{Late
  Start}, \emph{Early Stop}, \emph{Late Stop}, \emph{Surrounding}
(\emph{Early Start} \& \emph{Late stop}) and \emph{Consecutive} error.

Both \crf and \xlmr with projection settings have largely generated
\emph{Early Start} and \emph{Late Stop} errors. These models tend to
detect longer stimulus segments than annotated in the gold data. This
might be a consequence of English stimuli being longer than in
German. Despite the fact that a CRF does not have an understanding of
the length of span due to the Markov property, it has a bias weight
for transitions between I labels. An example for such a case is the
first instance from Table~\ref{tb:errors_exp} the projection setting
also extracted the token ``21-J\"ahriger'' as the start of the
stimulus sequence. This explains the difference between partial and
exact \F scores in Table~\ref{tb:crfs}.

The \emph{Surrounding} exemplifies that the models tend to predict the
beginning of a stimulus span directly after a colon.  In contrast, in
the in-corpus experiments (particularly with \xlmr), models tend to
generate \emph{Late Start} and \emph{Early Stop} errors more
often. For example the second headline from Table~\ref{tb:errors_exp}
shows a missing prediction of the verb ``scheitert''. Instead, the
preposition ``bei'' is found as the start of the stimulus
phrase. Further, in the subsequent example, this model setting does
not cover the phrase ``die ihr Kind mal gesagt hat'' in the stimulus
segment. Both sample headlines demonstrate that in-corpus models tend
to label prepositions as the start of stimulus sequences.

In the XML-R experiments, we opted against the use of a
Viterbi-decoded output layer (like a CRF output) -- this leads to
errors of the \emph{Consecutive} type, as shown in the last example:
start and end of the stimulus are correctly found, but tokens in
between have been missed.

\section{Conclusion and Future Work}

We introduced the first annotated German corpus for identifying
emotion stimuli and provided baseline model results for various CRF
configurations and an \xlmr model. We additionally
proposed a data projection method.

Our results show training and testing the model in the same language
outperforms cross-lingual models. Further, the \xlmr model that uses a
multilingual distributional semantic space outperforms the projection.
However, based on partial matches, we see that, when approximate
matches are sufficient projection and multilingual methods show an
acceptable result.

Previous work has shown that the task of stimulus detection can be formulated as
token sequence labeling or as clause classification \citep{Oberlaender2020}. In
this paper we limited our analysis and modeling on the sequence labeling
approach. Thus, we leave to future work the comparison with the
clause-classification approach. However, from the results obtained, we find
sequence labeling an adequate formulation in German.

For further future work, we suggest experimenting with the other existing
corpora in English to examine whether the cross-lingual approach would work
well on other domains. Regarding this, one could also train and improve models
not only for language change but also to extract stimuli across different
domains. Subsequently, another aspect that should be investigated is the
simultaneous recognition of emotion categories and stimuli.

\section*{Acknowledgements}
This work was supported by Deutsche Forschungsgemeinschaft (project
CEAT, KL 2869/1-2). Thanks to Pavlos Musenidis for fruitful
discussions and feedback on this study.

\bibliographystyle{acl_natbib}
\bibliography{lit}

\onecolumn
\appendix
\section{Appendix}

\begin{table*}[h]
  \centering
  \begin{tabular*}{\textwidth}{lll}
    \toprule
    Question & Annotation & Labels \\\cmidrule(r{5pt}){1-1} \cmidrule(l{5pt}r{5pt}){2-2} \cmidrule(l{5pt}r{-25pt}){3-3}
    \addlinespace
    \textbf{Phase 1:} Emotion Annotation & & \\
    \addlinespace
    1. Are there terms in the headline which could indicate an emotion? & Cue word & 0, 1  \\
    \addlinespace
    2. Does the text specify a person or entity experiencing an emotion? & Experiencer & 0, 1  \\
    \addlinespace
    3. Which emotion is most provoked within the headline? & Emotion & Emotions  \\\cmidrule(r{5pt}){1-1} \cmidrule(l{5pt}r{5pt}){2-2} \cmidrule(l{5pt}r{-25pt}){3-3}
    \addlinespace
    \textbf{Phase 2:} Stimuli & &  \\
    \addlinespace
    4. Which token sequence describes the trigger event of an emotion? & Stimulus & BIO  \\
    \bottomrule
  \end{tabular*}
  \caption{Questions for the annotation.}
  \label{tb:questionaires}
\end{table*}

\end{document}